\documentclass[10pt,twocolumn,letterpaper]{article}

\usepackage[normalem]{ulem}
\usepackage[pagenumbers]{cvpr} 

\usepackage[dvipsnames]{xcolor}

\definecolor{cvprblue}{rgb}{0.21,0.49,0.74}
\usepackage[pagebackref,breaklinks,colorlinks,citecolor=cvprblue]{hyperref}

\usepackage{soul}
\usepackage{amsmath,amssymb}
\usepackage[dvipsnames]{xcolor}
\usepackage[geometry]{ifsym}

\def\paperID{26} 
\def\confName{CVPR}
\def\confYear{2024}

\title{Segmentation of Maya hieroglyphs through fine-tuned foundation models}

\author{%
    FNU Shivam\textsuperscript{1},
    Megan Leight\textsuperscript{2},
    Mary Kate Kelly\textsuperscript{3},
    Claire Davis\textsuperscript{2},
    Kelsey Clodfelter\textsuperscript{2},
    Jacob Thrasher\textsuperscript{1}, \\
    Yenumula Reddy\textsuperscript{1},
    Prashnna Gyawali\textsuperscript{1}\\
    \textsuperscript{1} Lane Department of Computer Science and Electrical Engineering, West Virginia University, USA\\
    \textsuperscript{2} Department of Art History, School of Art and Design, West Virginia University, USA\\
    \textsuperscript{3} Mount Royal University, Canada\\
    {\tt\small \{ss00132,ced00010,kec00022,jdt0025\}@mix.wvu.edu, marykate.kelly89@gmail.com}, \\{\tt\small \{mleight,ramana.reddy,prashnna.gyawali\}@mail.wvu.edu}}

\begin{document}
\maketitle
\begin{abstract}
The study of Maya hieroglyphic writing unlocks the rich history of cultural and societal knowledge embedded within this ancient civilization's visual narrative. Artificial Intelligence (AI) offers a novel lens through which we can translate these inscriptions, with the potential to allow non-specialists access to reading these texts and to aid in the decipherment of those hieroglyphs which continue to elude comprehensive interpretation. Toward this, we leverage a foundational model to segment Maya hieroglyphs from an open-source digital library dedicated to Maya artifacts. Despite the initial promise of publicly available foundational segmentation models, their effectiveness in accurately segmenting Maya hieroglyphs was initially limited. Addressing this challenge, our study involved the meticulous curation of image and label pairs with the assistance of experts in Maya art and history, enabling the fine-tuning of these foundational models. This process significantly enhanced model performance, illustrating the potential of fine-tuning approaches and the value of our expanding dataset. We plan to open-source this dataset for encouraging future research, and eventually to help make the hieroglyphic texts legible to a broader community, particularly for Maya heritage community members.
\end{abstract}    
\section{Introduction}
\label{sec:intro}

\begin{figure*}[!t]
    \centering
    \includegraphics[scale=0.40]{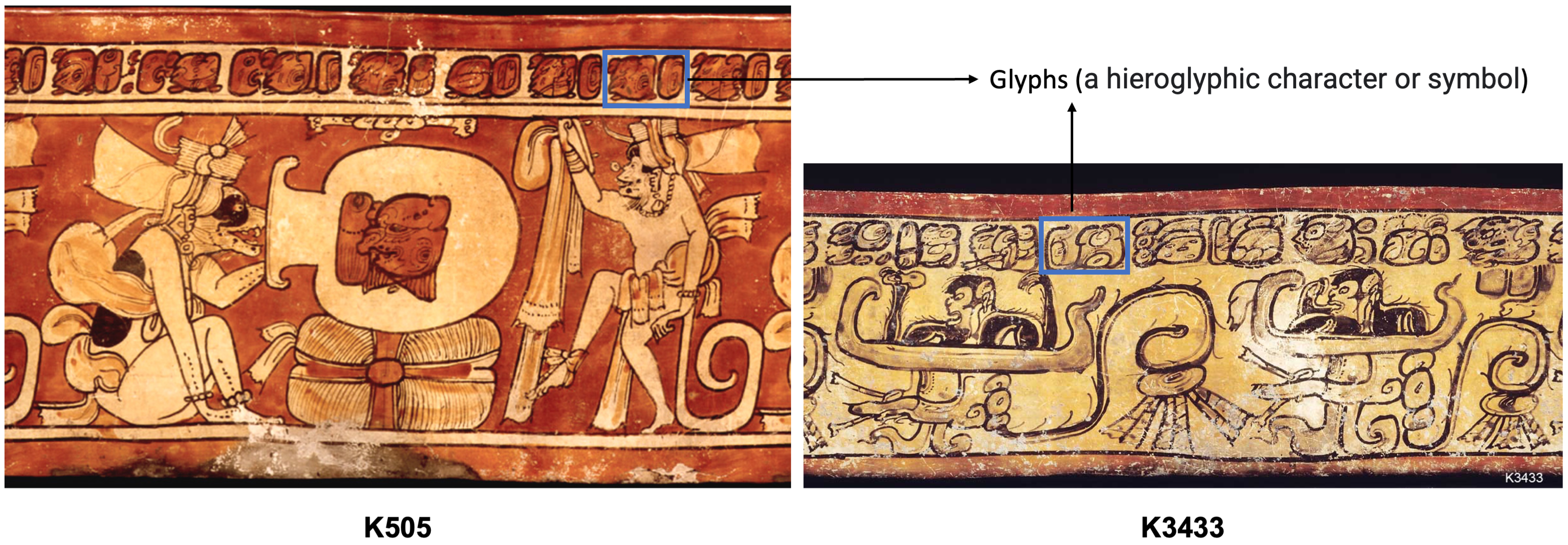}
    \caption{Examples of Maya vessels from Kerr's Maya Vase Database (K505 and K3433). A blue box highlights the shape of the glyph block, although K505 is pseudoglyphic. The segmentation of these glyph blocks, to their individual hieroglyphic characters or symbols, is the primary focus of this study.}
    \label{fig:imgExample}
\end{figure*}

The ancient Maya civilization, one of the most sophisticated and vibrant pre-Columbian societies in the Americas, has captivated scholars and enthusiasts alike with its rich tapestry of cultural achievements, particularly in art and writing \cite{fash1996building}. Renowned for their advanced knowledge in mathematics, astronomy, and architecture, the Maya archaeological record also includes a vast array of intricate art with hieroglyphic inscriptions that hold additional information to understanding their complex societal structures, history, rituals, and cosmology. Texts found on temple walls, stone monuments, pottery, and manuscripts all serve as a visual narrative of the ancient Maya people, offering invaluable insights into their daily lives, religious beliefs, and historical events. However, despite their significance, the intricate nature of the writing system is not easily parsed and presents a formidable challenge to traditional methods of study; it also serves as a barrier to interested non-specialists, and the modern Maya heritage community who are stakeholders in this historical knowledge.

Though the artifact record is robust, it is not nearly to the scale necessary for comprehensive decipherment of all known hieroglyphs. Many of them appear too infrequently in the corpus to have enough contexts to securely identify them. This is where artificial intelligence offers a groundbreaking opportunity in understanding Maya art, as demonstrated by previous research \cite{bogacz2018visualizing, hu2017extracting}. By leveraging AI and machine learning (ML) technologies, researchers can develop frameworks for the segmentation and interpretation of Maya hieroglyphs, thereby allowing for more comprehensive, thorough analysis of the corpus. Importantly, none of the existing AI frameworks are able to recognize nor segment the Maya glyph character blocks at the center of our study. Thus, this technological approach offers a promising solution to bridge the gap between the expert knowledge restricted to a small number of specialists and the interested public who could gain access to reading texts from this ancient culture. 

Current state-of-the-art AI frameworks like deep learning predominantly rely on extensive datasets, often in the magnitude of millions of well-annotated examples \cite{sun2017revisiting}. 
Such annotation is typically achieved through crowd-sourcing platforms, such as Mechanical Turk, where numerous individuals contribute to the labor-intensive process of tagging and categorizing data \cite{vaughan2018making}. This approach facilitates the development of AI systems capable of recognizing patterns and making informed predictions or analyses; however, the application of this methodology to the domain of ancient writing encounters significant obstacles.

The specialized nature of Maya hieroglyphs in ancient art means that accurately annotating these images requires a depth of expertise in the subject matter, a criterion that cannot be easily met through general crowd-sourcing methods. Experts in Maya epigraphy are relatively scarce, and the nuanced understanding required to identify and categorize the myriad of logograms (word signs) and syllabograms (sound signs) is not something that can be distilled into a simple task for the average person. Moreover, unlike the fields where AI has seen extensive application, the dataset of Maya artifacts is not vast but rather limited. Many of these artifacts are unique, with only a finite number available for study, further complicating the task of assembling a comprehensive and representative dataset for training AI models. This scarcity of data, coupled with the need for expert-level annotation, presents a substantial barrier to leveraging AI for the analysis and interpretation of Maya inscriptions, underscoring the need for innovative solutions tailored to the challenges of working with ancient and specialized datasets.

Foundational models are recent advancements within deep learning, and they are trained on vast datasets spanning millions of examples across various domains \cite{bommasani2021opportunities, kirillov2023segment}. These models have demonstrated remarkable capabilities, including zero-shot learning, which enables them to perform tasks on data they have never seen before, such as segmenting a wide array of images with precision, including complex medical images \cite{ma2024segment}. The success of foundational models in these areas, especially their ability to generalize and adapt to new challenges without direct training, has inspired us to explore their potential in the field of ancient writing.

Toward this end, we first explored the application of foundational models, specifically the Segment Anything (SAM) model \cite{kirillov2023segment}, for segmenting glyph blocks on Maya ceramic vessels. Despite its success in various domains, the performance of such foundational segmentation models was initially sub-optimal in this context. Secondly, recognizing the critical importance of expertly annotated data in addressing the challenges posed by the limited and specialized nature of Maya hieroglyphs, we enlisted the help of several individuals with deep expertise in Maya art and epigraphy. These experts dedicated their efforts to meticulously annotating images sourced from Justin Kerr's Maya Vase Database \cite{kerr_maya_vase}, which serves as a photographic repository for ancient Maya artifacts. In focusing on the vase database, each vessel was carefully annotated around each glyph block to ensure the highest level of accuracy. Finally, we fine-tuned the SAM model using our specifically curated dataset, which led to a demonstrable improvement in performance. Overall, our contributions can be summarized as follows:

\begin{itemize}
    \item For the first time, we explored the capabilities of publicly available foundational models, specifically Segment Anything (SAM), in segmenting hieroglyphs from Maya vessels.
    \item Our study involved the manual annotation of hieroglyphs within publicly available images of Maya art, which we plan to open source to empower other researchers with a valuable resource for exploring and understanding Maya hieroglyphs.
    \item We fine-tuned SAM, demonstrating improved segmentation performance.
\end{itemize}

\section{Dataset preparation and Pre-processing}

\subsection{Data Sources}
The dataset utilized in our project is derived from the MayaVase database, an extensive collection of images curated by Justin Kerr from 1966-2012, recently acquired by Dumbarton Oaks Research Library and Collection \cite{kerr_maya_archive}. The database comprises images of vessels and objects housed in museums and private collections, the majority of which have been gathered from across the United States, Europe, Mexico, and Central America. The MayaVase database is particularly notable for its publication of Kerr's pioneering rollout photographs of Maya vessels, which allowed cylindrical objects to be viewed as a single narrative image \cite{kerr2007short}. In addition to cylindrical vessels, the database includes images of plates, bowls, archaeological sites, various art objects, and items from private collections, providing a comprehensive visual catalog of Maya material culture in the public domain. We have provided two examples of vessels from the Maya Vase Database in Fig. \ref{fig:imgExample}. Note the presence of pseudoglyphic text on K505, the vessel with the dog peering into a vase. The text is placed correctly and the blocks are glyph-shaped; however, they do not bear legible meaning, which confirms the importance of having Maya epigraphers on these types of AI data research projects.

\subsection{Data Preparation}
Annotation of the images represented a pivotal contribution to our project, encompassing extensive efforts in data preparation by a dedicated team comprising faculty and students from art historical and anthropological disciplines. These team members, who have devoted years to the study of these objects, played a crucial role in the meticulous process of data curation and annotation. The process began with a comprehensive web-scraping of the entire Maya Vase Database to collect our sample images. Kerr, in collaboration with other specialists, analyzed each vessel and assigned similar iconographic themes (e.g. bees, canoe, dog) across all the images in the database. We focused on vessels from approximately 27 of the 88 different iconographic themes.

\begin{figure}[!t]
  \centering
  \begin{subfigure}{0.5\linewidth}
    \centering
    \includegraphics[width=\linewidth]{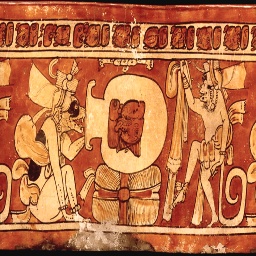}
    \caption{Input image}
    \label{fig:sub1}
  \end{subfigure}%
  \begin{subfigure}{0.5\linewidth}
    \centering
    \includegraphics[width=\linewidth]{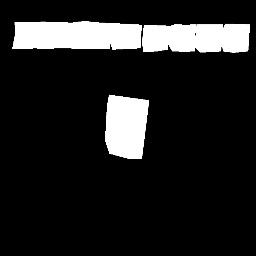}
    \caption{Binary mask}
    \label{fig:sub2}
  \end{subfigure}
  \caption{Binary mask (right) generated through manual annotation of glyphs from the original input image (left).}
  \label{fig:sampleImgMask}
\end{figure}

Subsequent to the image collection, the team embarked on the manual annotation of each image, carefully selecting those containing text. Hieroglyphs were present in each image, resulting in a curated collection of over 117 images, varying in shape and size, specifically chosen for their presence of glyph blocks or running hieroglyphic text. Then, more detailed annotation took place by carefully identifying and labeling each block, a task that required a deep understanding of Maya iconography and script. To facilitate this process, we utilized labelme.com, an open-source annotation tool, which allowed for efficient and accurate annotation work. For each image, polygons were drawn around each hieroglyph block and given different file names, and a corresponding JSON file was created, containing all the annotations for a single image. These JSON files were used in generating ground truth binary masks for each image. An example of input image and its corresponding binary mask is presented in Fig. \ref{fig:sampleImgMask}.

\section{Methodology}
We present a novel approach for segmenting Maya characters using a foundational segmentation model. Specifically, we consider the Segment Anything (SAM) \cite{kirillov2023segment} model and fine-tuned this model with our dataset.

\subsection{Preliminary: Segment Anything (SAM)}
The Segment Anything (SAM) model is a foundational segmentation network, comprising an image encoder, a prompt encoder, and a mask decoder \cite{kirillov2023segment}. 
SAM works by first encoding the image into a high dimensional vector representation using the image encoder, which is a Masked Autoencoder pre-trained Vision Transformer. The prompt is encoded into a separate vector representation by the prompt encoder. These two vector representations are then combined and passed to a mask decoder, which gives the output specified by the prompt.

Even though SAM has shown great promise in segmenting objects that it has never seen before, there are several issues when it is applied to Maya hieroglyph segmentation. Maya images have too much noise and unwanted characters, which makes it difficult for SAM to segment the Maya hieroglyphs from the other elements of the visual scenes. Despite the fact that SAM is trained on using a large and diverse number of images, they are not specific to the ancient Maya image datasets, which makes segmentation of Maya characters more inaccurate than what we expected. So, in this regard, fine-tuning the SAM on these ancient Maya images is important to move forward.

\subsection{Finetuned-SAM}
We propose fine-tuning SAM to enhance segmentation performance for Maya hieroglyph recognition. By refining SAM, we aim to adapt it specifically to the nuances of Maya inscriptions. For fine-tuning, we utilize the pre-trained weights of the image encoder and prompt encoder, ensuring the model benefits from prior learning. However, we freeze the parameters of the image and prompt encoder, calculating gradients only for the mask decoder.

Towards learning the Finetuned-SAM, we use the combination of Dice loss and Cross-Entropy loss as the objective function. The overall loss can be represented as: 
$$L = \alpha \cdot L_{\text{CE}} + \beta \cdot L_{\text{Dice}}$$
where $\alpha$ and $\beta$ are the weights that determine the contribution of each loss component to the total loss. In our experiments, both components are given equal importance.

\begin{figure}[!t]
    \centering
    \includegraphics[scale=0.57]{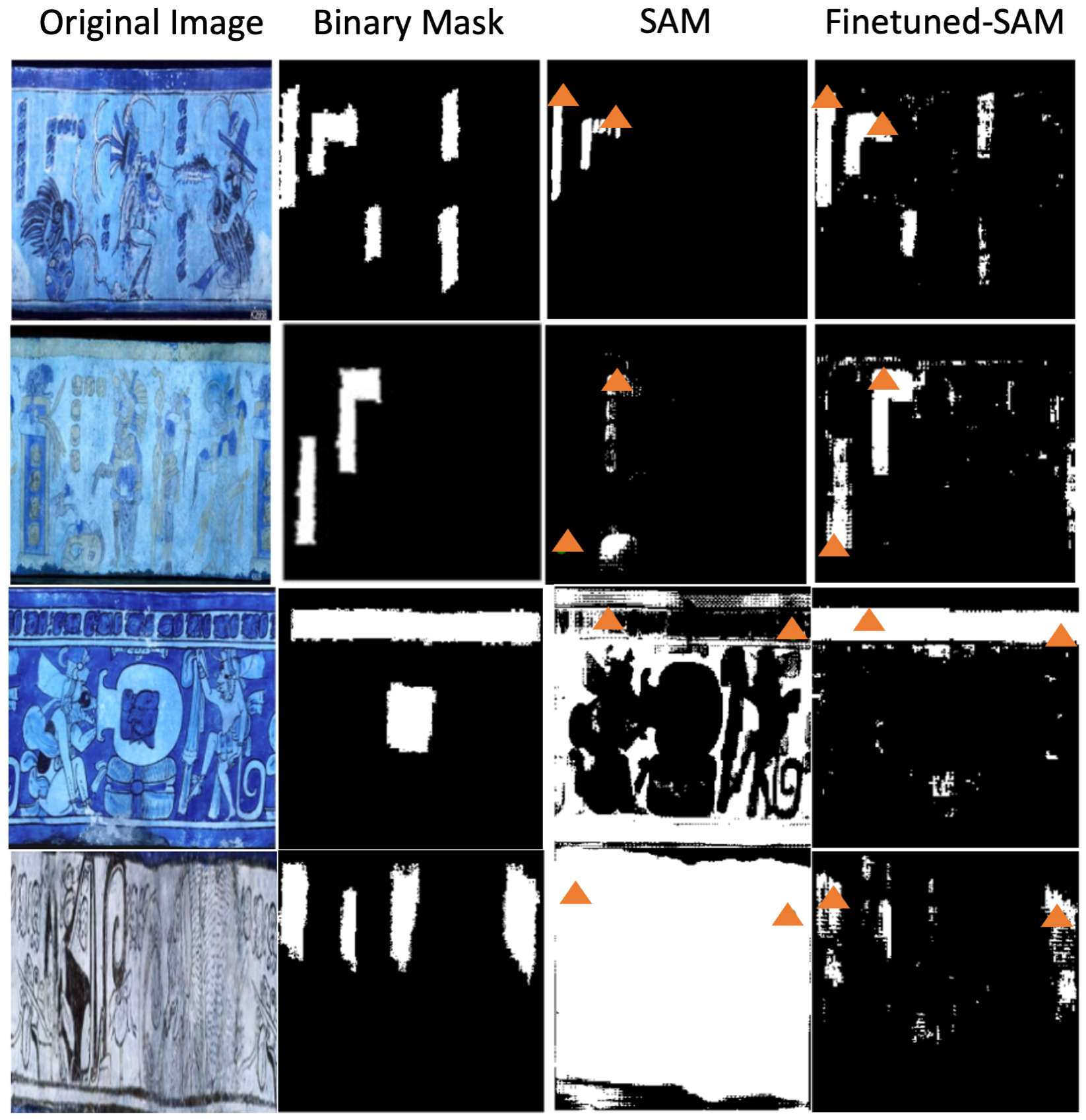}
    \caption{Qualitative analysis of the results from SAM and Finetuned-SAM for selected samples in our test dataset. For both SAM and Finetuned-SAM, we display the result when two random points are chosen as prompts (denoted by {\color{Orange}\FilledBigTriangleUp}).}
    \label{fig:qualitative1}
\end{figure}

\section{Experiments and Results}

\subsection{Setup}

{\bf Data:} Our experiment utilized a dataset of 117 total images with corresponding annotations for the Maya hieroglyphs in each image, generated in the form of a binary mask. 
Each image and its corresponding ground truth mask was resized to 256 by 256 pixels in order to standardize the input size. 
This dataset is partitioned randomly into three subsets: training (75 images), validation (19 images), and test (23 images).
All the reported results, including evaluation metrics, are based on the test split.

\noindent{{\bf Model:}} For modeling, we utilized the SAM ViT Base version of the SAM model. For fine-tuning the SAM model (Finetuned-SAM), we trained with the following hyperparameters: 300 epochs, a learning rate of 1e-3, a batch size of 6, and used the Adam optimizer. A validation split was employed to identify the best-performing model during training. We tracked the model's performance on the validation set and selected the model demonstrating the highest performance for testing. The compute time on a NVIDIA RTX 3090 24 GB for fine-tuning was approximately 2 hours.
We assessed our model's performance using two standard metrics: Intersection-over-Union (IoU) and DICE score.

\subsection{Results}
We provide the main results of our study in Table \ref{tab:table1}. As shown, we also compared our models against standard segmentation approaches like UNet \cite{ronneberger2015u} and autoencoder-based segmentation \cite{ohgushi2020road,li2019deep}. Compared to these, the foundation-based approach demonstrates improved performance. Within the foundation-based approach, our proposed Finetuned-SAM model outperforms the off-the-shelf SAM model. Given that these foundation models rely on prompts, we explored various prompt options in our inference analysis. Additionally, to ensure unbiased results, we conducted our inference analysis five times with randomly selected prompts.

\begin{table}[h]
  \centering
  \begin{tabular}{@{}lcc@{}}
    \toprule
    Model & IoU & DICE   \\
    \midrule
    UNet & 0.179 & 0.289 \\
    Autoencoder & 0.174 & 0.277  \\
    \hline 
    SAM [1 random point] & 0.118 & 0.199  \\
    SAM [2 random point] & 0.115 & 0.196  \\
    SAM [0.5 Bbox] & 0.128 & 0.219  \\
    SAM [0.75 Bbox] & 0.097& 0.173\\
    \hline
    Finetuned-SAM [1 random point] & 0.395 & 0.532  \\
    Finetuned-SAM [2 random point] & 0.397 & 0.538  \\
    Finetuned-SAM [0.5 Bbox] & 0.327 & 0.471  \\
    Finetuned-SAM [0.75 Bbox] & 0.321 & 0.453 \\
    \bottomrule
  \end{tabular}
  \caption{The comparison of the presented Finetuned-SAM with standard segmentation methods and regular SAM evaluated by IoU and Dice score. For SAM and Finetuned-SAM, we report the average of five random inference runs. }
  \label{tab:table1}
\end{table}


\begin{table}[t]
\small
\centering

\begin{tabular}{@{}lcc@{}}
\toprule
Model & IoU & DICE \\
\midrule
SAM [1 random point/block] & 0.316 & 0.438 \\
SAM [2 random points/block] & 0.387 & 0.522 \\
SAM [3 random points/block] & 0.382 & 0.517 \\
\midrule
Finetuned-SAM [1 random point/block] & 0.417 & 0.555 \\
Finetuned-SAM [2 random points/block] & 0.439 & 0.580 \\
Finetuned-SAM [3 random points/block] & 0.410 & 0.543 \\
\bottomrule
\end{tabular}

\caption{ Quantitative analysis of Maya glyph segmentation with individualized input prompts demonstrating improved accuracy.}
\label{tab:table2}
\end{table}

Further, we analyzed the effect of providing prompts for individual glyph blocks in our database. We present the results in Table \ref{tab:table2}, where compared to Table \ref{tab:table1}, both SAM and Finetuned-SAM's performances have improved. This improvement is anticipated, as providing block-specific prompts offers more spatial context. Nonetheless, even in this scenario, Finetuned-SAM outperforms SAM, demonstrating the effectiveness of fine-tuning.


In Figure \ref{fig:qualitative1}, we provide qualitative results comparing the performance of SAM and Finetuned-SAM. As observed, Finetuned-SAM outperforms SAM in segmenting hieroglyphs from Maya images across all examples. Notably, in some cases (third and fourth row), the SAM model fails significantly, whereas Finetuned-SAM, though not perfect, still focuses on the correct region.




\subsection{Conclusion}
We present Maya hieroglyph segmentation using a foundational segmentation model, which could lay the groundwork for using AI in the recognition of these hieroglyphs. We employed the foundational segmentation model, Segment Anything (SAM), as a baseline to segment Maya texts, and introduced Finetuned-SAM, a version of the SAM model fine-tuned with a dataset meticulously annotated with the assistance of experts in Maya art and writing. The results from the Finetuned-SAM model underscore its superior ability to segment inscriptions accurately, even when training is conducted with a relatively limited dataset. Our immediate future work includes expanding the dataset, as it is evident that a more annotated dataset would potentially improve the results further. The public release of these data could potentially ignite interest among researchers in using recent advancements in AI to study ancient Maya writing, as well as provide access to the inscriptions and their translations to interested publics - both for the Maya heritage community to gain increased access to their ancestral textual records, as well as for interested non-specialists.

\subsection{Acknowledgement}
We are grateful to the Humanities Center at West Virginia University for their generous support of this research through a humanities collaboration grant, which will enhance our project and its contributions to the academic community.  

{
    \small
    \bibliographystyle{ieeenat_fullname}
    \bibliography{main}
}


\end{document}